\documentclass[11pt]{article}

\usepackage[final]{acl}

\usepackage{times}
\usepackage{latexsym}
\usepackage{epigraph}

\usepackage[T1]{fontenc}
\usepackage[utf8]{inputenc}

\usepackage{microtype}

\usepackage{inconsolata}

\usepackage{graphicx}

\title{The Rashomon Wikipedia: A Data-Perspectivist Analysis of Divergent Historical Narratives}

\author{
Claudiu Creanga$^{2, 3}$, Liviu P. Dinu$^{1, 3}$, Anca Dinu$^{3, 4}$\\ 
  $^1$ Faculty of Mathematics and Computer Science, \\
  $^2$ Interdisciplinary School of Doctoral Studies, \\
  $^3$ HLT Research Center, \\
  $^4$ Faculty of Foreign Languages and Literatures, \\
  University of Bucharest, Romania\\
  \small
{\tt claudiu.creanga@fmi.unibuc.ro, ldinu@fmi.unibuc.ro, anca.dinu@lls.unibuc.ro}  \\
}

\begin{document}
\maketitle
\begin{abstract}
  Wikipedia aims to provide a unified, neutral record of history, yet its independent language editions often function as distinct epistemic communities, creating divergent narratives around contested events. This paper investigates cross-lingual historiographical bias by analyzing Wikipedia articles across five languages (Romanian, Hungarian, Russian, Turkish, and English) focusing on three contentious events in Romanian history: the Battle of Posada (1330), the Soviet occupation of Bessarabia (1940), and the Night Attack at Târgoviște (1462). Using human annotators and Large Language Models (LLMs) to classify citation stance and quantify narrative evolution from 2005 to 2024, we identify a phenomenon of "citation isolation". In the case of the Battle of Posada, only 2 out of 119 citations were shared between language editions, with the Romanian edition exhibiting a 91\% pro-national bias compared to the balanced Hungarian edition. Longitudinal analysis reveals that these narratives are volatile and responsive to contemporary geopolitics, evidenced by a significant shift in the Russian framing of Bessarabia in 2024. Finally, we propose a "Peace-Maker" pipeline to automate conflict reconciliation. We demonstrate that while standard prompting leads models to hallucinate consensus, "adversarial" prompting, which explicitly instructs the model to preserve and attribute disagreement, achieves near-perfect neutrality scores.
\end{abstract}

\epigraph{History is that certainty produced at the point where the imperfections of memory meet the inadequacies of documentation.}{Julian Barnes}

\section{Introduction}

Wikipedia is the first place many people turn to for historical information. It aims to provide a "Neutral Point of View" (NPOV), representing conflicting perspectives fairly and without bias. However, Wikipedia is not a single unified record. It is a federation of independent language editions, each written by a distinct community of editors who often rely on their own national historiography. When history is contested, such as in wars or territorial disputes, these communities can produce articles that describe the same event in fundamentally different ways.

This problem goes beyond simple translation differences. A reader of the Romanian Wikipedia might learn about a heroic defense of independence, while a reader of the Turkish or Russian Wikipedia reads about a failed rebellion or a justified annexation. While the NPOV policy exists in all editions, its application varies. Local editors often cite local sources, creating "citation universes" that rarely overlap. As a result, the bias is not just in the text, but in the choice of which facts to include and which authorities to trust.

In this paper, we study how these conflicting narratives form and evolve. We focus on three events that are central to Romanian history but contested by its neighbors: the Battle of Posada (1330) against Hungary, the Soviet occupation of Bessarabia (1940) involving Russia, and the Night Attack at Târgoviște (1462) against the Ottoman Empire. We analyze articles from five language editions (Romanian, Hungarian, Russian, Turkish, and English) to answer three questions:

\textbf{1. How do citation choices reflect national bias?} We find that editors almost exclusively cite authors from their own country. In our analysis of the Battle of Posada, only 2 out of 119 citations were shared between the Romanian, Hungarian, and English editions.

\textbf{2. How do narratives evolve over time?} By analyzing article versions from 2005 to 2024, we show that these narratives are not static. While the Romanian account of the Night Attack has moderated in recent years, the Russian account of Bessarabia has been volatile, shifting significantly in 2024.

\textbf{3. Can AI help reconcile these conflicts?} We test whether LLMs can act as neutral arbiters. We find that standard prompting fails because models try to "fix" the conflict by choosing one side. However, an "adversarial" prompting strategy, which explicitly instructs the model to preserve the disagreement, can generate summaries that are rated as perfectly neutral.

It is important to note that "neutrality" in history is a contested concept. Often, there is no single objective truth between two national myths. Therefore, we do not define neutrality as finding a "middle ground" or a "correct" version of events. Instead, we define it operationally as \textit{perspectival balance}: a neutral summary is one that accurately represents the existence and nature of the conflict itself, attributing claims to their respective traditions without endorsing one over the other.

\section{Related Work}
\label{sec:related_work}

The challenge of addressing conflicting historical narratives on Wikipedia sits at the intersection of computational social science, natural language processing, and digital history. While early work focused on metadata-driven analysis of "edit wars", recent advances in LLMs have shifted attention toward semantic analysis of bias and automated conflict resolution.

\subsection{Epistemic Communities \& Cross-Lingual Divergence}
Wikipedia is ostensibly a global project, but empirical research suggests it functions more as a federation of distinct "epistemic communities" \citep{samoilenko2016linguistic}. \citet{hecht2009measuring} first quantified the "self-focus bias" inherent in these communities, showing that language editions disproportionately cover topics related to their own geography. \citet{miquel2018wikipedia} expanded this to 40 languages, revealing a persistent "culture gap" where shared knowledge is the exception rather than the rule.

In the domain of history, this divergence often manifests as "citation isolation". \citet{zheng2023evaluating} found that non-English Wikipedias cite millions of unique sources not found in English Wikipedia, effectively creating parallel knowledge bases. \citet{baigutanova2023reference} further showed that sources deemed unreliable in one language often persist in others, highlighting the lack of a unified standard for verifiability. Our work provides a granular case study of this phenomenon: we show that for the Battle of Posada, the "Romanian epistemic community" and the "Hungarian epistemic community" rely on entirely disjoint sets of historical authorities, with only 2 shared sources out of 119.

\subsection{Bias Detection: From Syntax to Semantics}
Traditional approaches to bias detection in NLP have focused on linguistic cues. \citet{recasens2013linguistic} created the foundational "NPOV corpus", identifying subjective intensifiers (e.g., "famous", "outrageous") as markers of bias. \citet{pryzant2020neutralizing} advanced this by using BERT-based models to automatically rewrite such sentences into neutral forms.

However, historical bias is often implicit and structural rather than purely stylistic. \citet{Rogers2012} demonstrated this in their manual analysis of the Srebrenica massacre articles, where the bias lay in \textbf{which} facts were selected rather than \textbf{how} they were phrased. Recent work by \citet{narrative_bias_2025} confirms that modern LLMs still struggle with this "narrative bias", often reproducing intersectional identity biases even when explicitly instructed to be neutral. Our methodology addresses this by moving beyond sentence-level style transfer to document-level narrative analysis, using LLMs to quantify the "stance" of citations and the evolution of historical framing over decades.

\subsection{Perspectivism and Automated Reconciliation}
The emerging "Perspectivist Data" paradigm \citep{perspectivist_manifesto} argues that disagreement in data should not always be aggregated away or treated as noise. Instead, valid conflicting perspectives should be preserved. This is particularly relevant for LLMs, which tend to hallucinate a single "consensus" reality when none exists. \citet{kaneko2023language} and \citet{geopolitical2023} have shown that LLMs exhibit strong "nationality bias", often aligning with the geopolitical views of their training data's dominant language.

To mitigate this, recent frameworks like "MoDS" (Moderating a Mixture of Document Speakers) \citep{mods2025} and "Multi-Perspective Fusion" \citep{mpf_2025} have proposed treating documents as debating agents. Our "Peace-Maker" pipeline extends this line of inquiry. We show that standard "summarization" prompts fail because they encourage the model to resolve conflict (a form of \textbf{hallucinated consensus}), whereas "adversarial" prompts that enforce the preservation of conflict, aligning with the perspectivist approach, achieve significantly higher neutrality scores.

\section{Methodology}
\label{sec:methodology}

Our approach combines historical data mining with LLM-based analysis to quantify bias and generate neutral narratives. The pipeline consists of three stages: (1) multi-lingual data collection, (2) granular stance classification, and (3) adversarial narrative generation.

\subsection{Data Collection}
We targeted three historical events chosen for their contentious nature in Eastern European historiography:
\begin{enumerate}
    \item \textbf{Battle of Posada (1330):} A foundational conflict between Wallachia and Hungary.
    \item \textbf{Soviet Occupation of Bessarabia (1940):} A territorial dispute between Romania and the USSR/Russia.
    \item \textbf{Night Attack at Târgoviște (1462):} A military encounter between Vlad the Impaler (Wallachia) and Mehmed II (Ottoman Empire).
\end{enumerate}

For the \textbf{citation analysis} (Posada), we extracted the full content of the Romanian (ro), Hungarian (hu), and English (en) Wikipedia articles as of February 2026. We used a standard github library \citep{mwparserfromhell} to parse the MediaWiki markup and extract all citations, including those in `<ref>` tags and bibliography sections.

For the \textbf{temporal analysis} (Bessarabia and Târgoviște), we used the Wikipedia Action API \citep{wikipedia_api} to fetch historical snapshots of the articles from January 1st of 2005, 2010, 2015, 2020, and 2024. This resulted in a dataset of 29 article versions across Romanian, Russian, Turkish, and English editions.

\subsection{Stance Classification \& Metrics}
To quantify bias, we used LLMs and human annotators.

\subsubsection{Citation Stance (Posada)}
We classified 119 citations into four categories: \textit{Pro-Romanian}, \textit{Pro-Hungarian}, \textit{Neutral}, and \textit{Contested}. 
A major challenge was that many citations in the bibliography lacked context (e.g., just "Djuvara, p. 180"). To address this, we developed a \textbf{Context Enhancer} module. For each citation, the module searched the full article text for mentions of the author or title, extracting a $\pm$300 character window around the mention. This "enhanced context" was then fed to Google's Gemini 3.0 Pro Preview model \citep{geminiteam2025gemini} with the following prompt structure:
\begin{quote}
    \small "You are an expert historian analyzing the Battle of Posada (1330) between Wallachia and Hungary. Analyze the following text snippet from a Wikipedia article to determine the STANCE of the citation. The citation being analyzed is: "${citation-text}$". The surrounding context is: "${context}$". Determine if the citation is used to support a specific narrative:: Pro-RO (heroic defense), Pro-HU (treacherous trap), Neutral, or Contested".
\end{quote}
All API calls were executed with a temperature of 0 to ensure deterministic reproducibility.

\subsubsection{Human validation}
To validate the LLM's classifications, we used two human annotators. One annotator reviewed a random 50\% sample of the Romanian and English citations, while a second annotator reviewed the complete set of Hungarian citations. Both annotators assessed the same "enhanced context" provided to the LLM using the identical 4-category scheme (Pro-RO, Pro-HU, Neutral, Contested), blinded to the LLM's ratings. The human-LLM agreement was substantial, yielding Cohen's Kappa scores of 0.80 for Romanian, 0.75 for English, and 0.85 for Hungarian.

\subsubsection{Granular Narrative Metrics (Bessarabia/Târgoviște)}
For the temporal analysis, we defined event-specific metrics on a 0-100 scale. For example, for the Night Attack, we measured:
\begin{itemize}
    \item \textbf{Vlad Heroism:} Extent to which Vlad is portrayed as a brave defender.
    \item \textbf{Ottoman Threat:} Emphasis on the magnitude of the invading force.
    \item \textbf{Cruelty:} Emphasis on impalement and brutal tactics.
\end{itemize}
We processed each article version through the LLM to score these metrics, allowing us to track the evolution of the narrative over 20 years. The full prompt used for this granular scoring is provided in Appendix \ref{app:prompt}.

While the prompt requested categorical stances and confidence scores (0.0-1.0), we aggregated these into 0-100 scales for visualization by mapping the confidence of a "Pro-X" classification to a positive score and "Pro-Y" to a negative or lower score, normalized to a 0-100 range where 100 represents the maximum intensity of the national narrative.

\subsection{The "Peace-Maker" Pipeline}
To address the challenge of representing conflicting narratives, we developed the "Peace-Maker" pipeline.

\begin{enumerate}
    \item \textbf{Claim Extraction:} We first extract key factual claims from each source text (e.g., "Romanian source claims 15,000 Ottoman casualties").
    \item \textbf{Conflict Matching:} We use an LLM to identify pairs of conflicting claims (e.g., "RO: Attack was a victory" vs. "TR: Attack was a failed assassination").
    \item \textbf{Adversarial Generation:} We tested four prompting strategies to generate a summary:
    \begin{itemize}
        \item \textit{Standard:} "Summarize these sources".
        \item \textit{Academic:} "Write as a peer-reviewed historian".
        \item \textit{Mediator:} "Write a diplomatic report acceptable to both sides".
        \item \textit{Adversarial:} "Preserve the conflict. Explicitly state 'Source A says X while Source B says Y'. Do not resolve the dispute".
    \end{itemize}
    \item \textbf{LLM-as-a-Judge:} We evaluated the generated summaries using a separate LLM instance to score them on \textit{Neutrality} (0-1), \textit{Conflict Preservation} (0-1), and \textit{Source Attribution} (0-1).
\end{enumerate}

\section{Experiments \& Results}
\label{sec:experiments}

We present results from three experiments: (1) citation stance analysis of the Battle of Posada, (2) temporal analysis of narratives surrounding Bessarabia and Târgoviște, and (3) the evaluation of the Peace-Maker pipeline.

\subsection{Experiment 1: Citation Isolation (Posada)}
We analyzed 119 citations across the Romanian (RO), Hungarian (HU), and English (EN) Wikipedia articles on the Battle of Posada (1330).

\begin{figure}[h]
    \centering
    \includegraphics[width=\columnwidth]{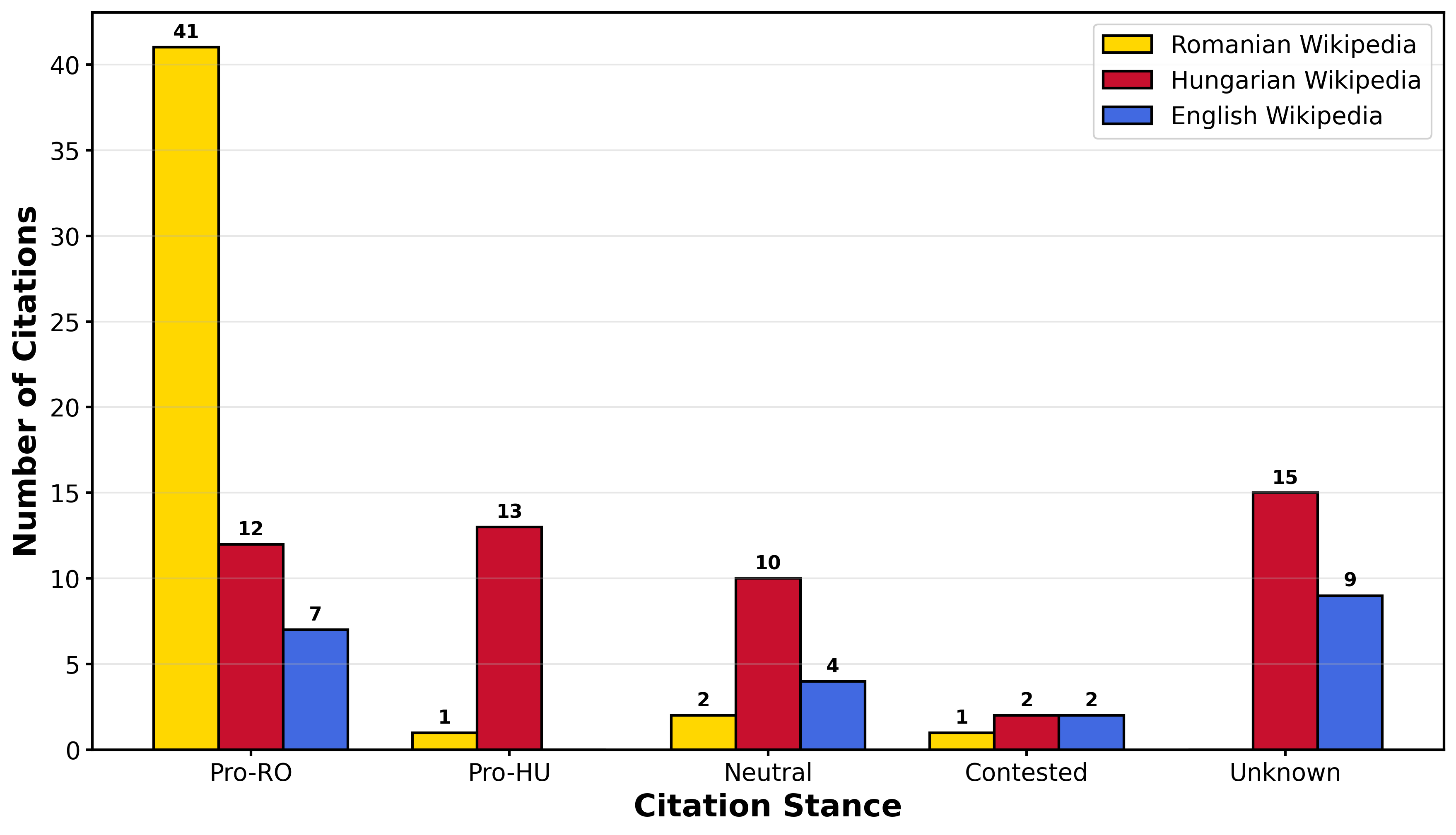}
    \caption{Raw count of citations by stance. Note the overwhelming volume of Pro-Romanian citations in the Romanian edition compared to the balanced distribution in Hungarian.}
    \label{fig:posada_dist}
\end{figure}

\begin{figure}[h]
    \centering
    \includegraphics[width=\columnwidth]{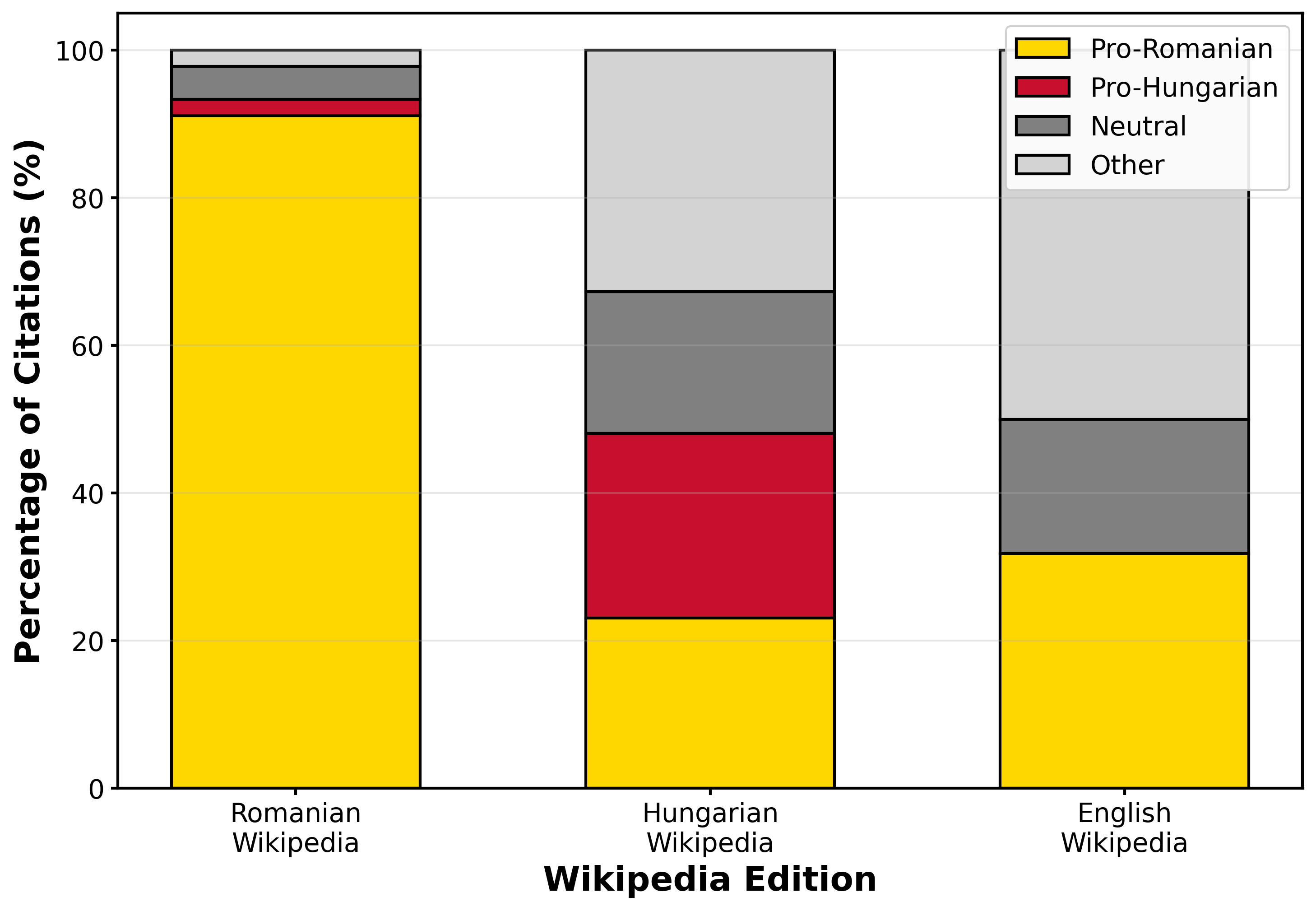}
    \caption{Stance distribution of citations in the Battle of Posada articles. Romanian Wikipedia is 91\% Pro-Romanian, while Hungarian Wikipedia is balanced.}
    \label{fig:posada_stance}
\end{figure}

\textbf{Bias Scores:} The Romanian edition exhibited a strong national bias, with 91.1\% of citations classified as \textit{Pro-Romanian} and only 2.2\% as \textit{Pro-Hungarian} (Figure \ref{fig:posada_stance}). In contrast, the Hungarian edition was remarkably balanced, with 23.1\% \textit{Pro-Romanian} and 25.0\% \textit{Pro-Hungarian} citations. The English edition, often assumed to be neutral, showed a moderate Pro-Romanian lean (31.8\% Pro-RO vs 0\% Pro-HU).

Figure \ref{fig:posada_dist} highlights the sheer volume disparity. It is important to note that raw citation volume is not a proxy for quality; a higher count could simply indicate a more detailed article. However, in this context, the volume reinforces the echo chamber. The Romanian article builds an "illusion of consensus" through the repetition of a large, mono-perspectival corpus, whereas the Hungarian article achieves a balanced narrative with a smaller but more diverse set of references.

\begin{figure}[h]
    \centering
    \includegraphics[width=\columnwidth]{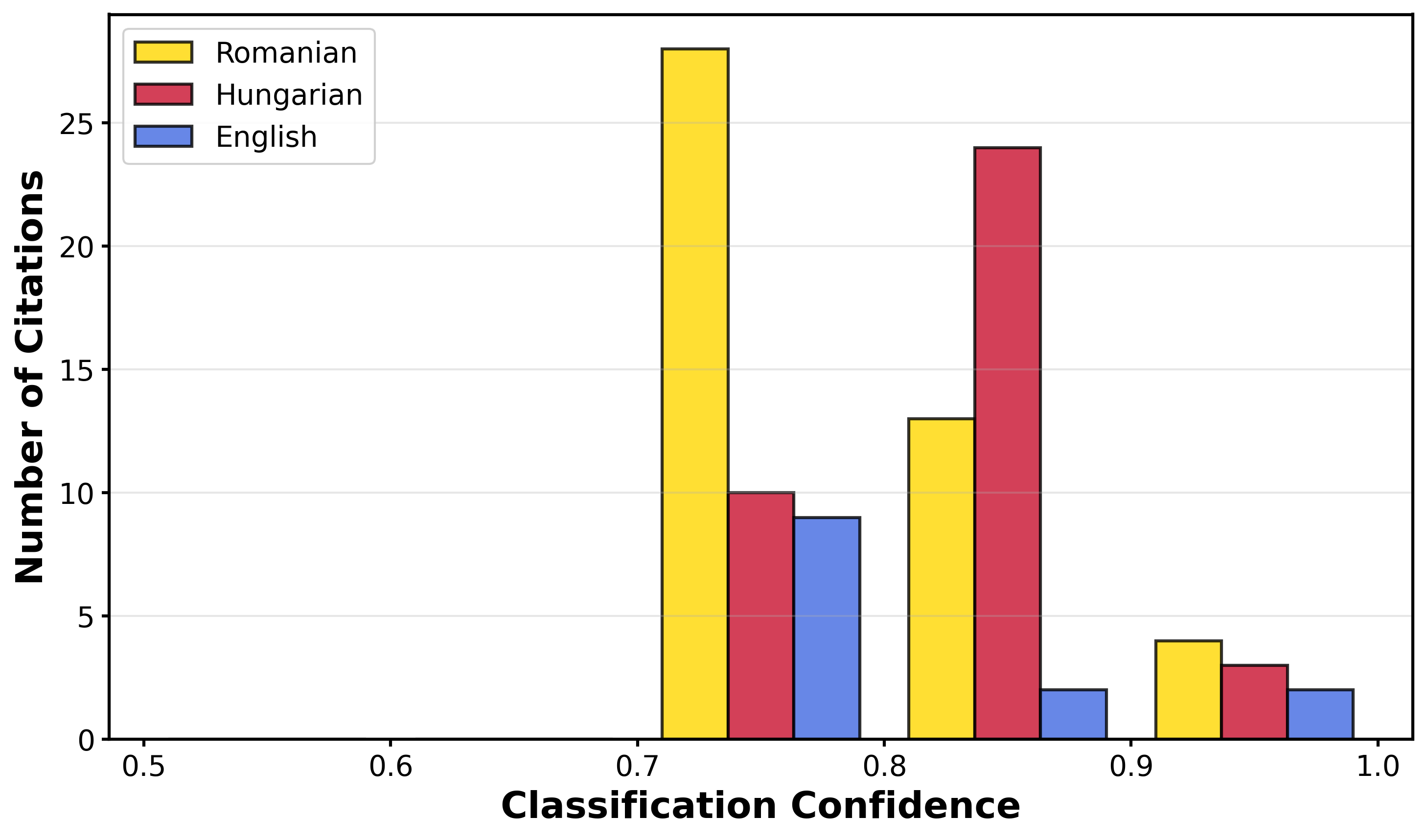}
    \caption{LLM Confidence Distribution. The high confidence scores across all stances indicate that the "Citation Isolation" is not due to model uncertainty but reflects clear, distinct narrative framing in the sources.}
    \label{fig:posada_conf}
\end{figure}

\textbf{Citation Isolation:} The most striking finding was the lack of overlap. Out of 119 unique citations, only \textbf{2 sources} were shared across more than one language edition. This confirms that these Wikipedia communities operate in distinct "citation universes", constructing their narratives from entirely disjoint sets of historical authorities. It is important to note that this isolation is partly driven by linguistic accessibility: Romanian scholars naturally cite Romanian historians writing in Romanian, while Hungarian scholars cite Hungarian sources. However, the result is the same: readers of different language editions are presented with fundamentally different evidentiary bases. As shown in Figure \ref{fig:posada_conf}, the LLM's classification confidence remained high across all categories, validating that these sources express clear, unambiguous stances rather than vague or subtle biases.

\subsection{Experiment 2: Temporal Evolution}
We tracked narrative metrics across 29 article versions from 2005 to 2024.

\begin{figure*}[t]
    \centering
    \includegraphics[width=0.48\textwidth]{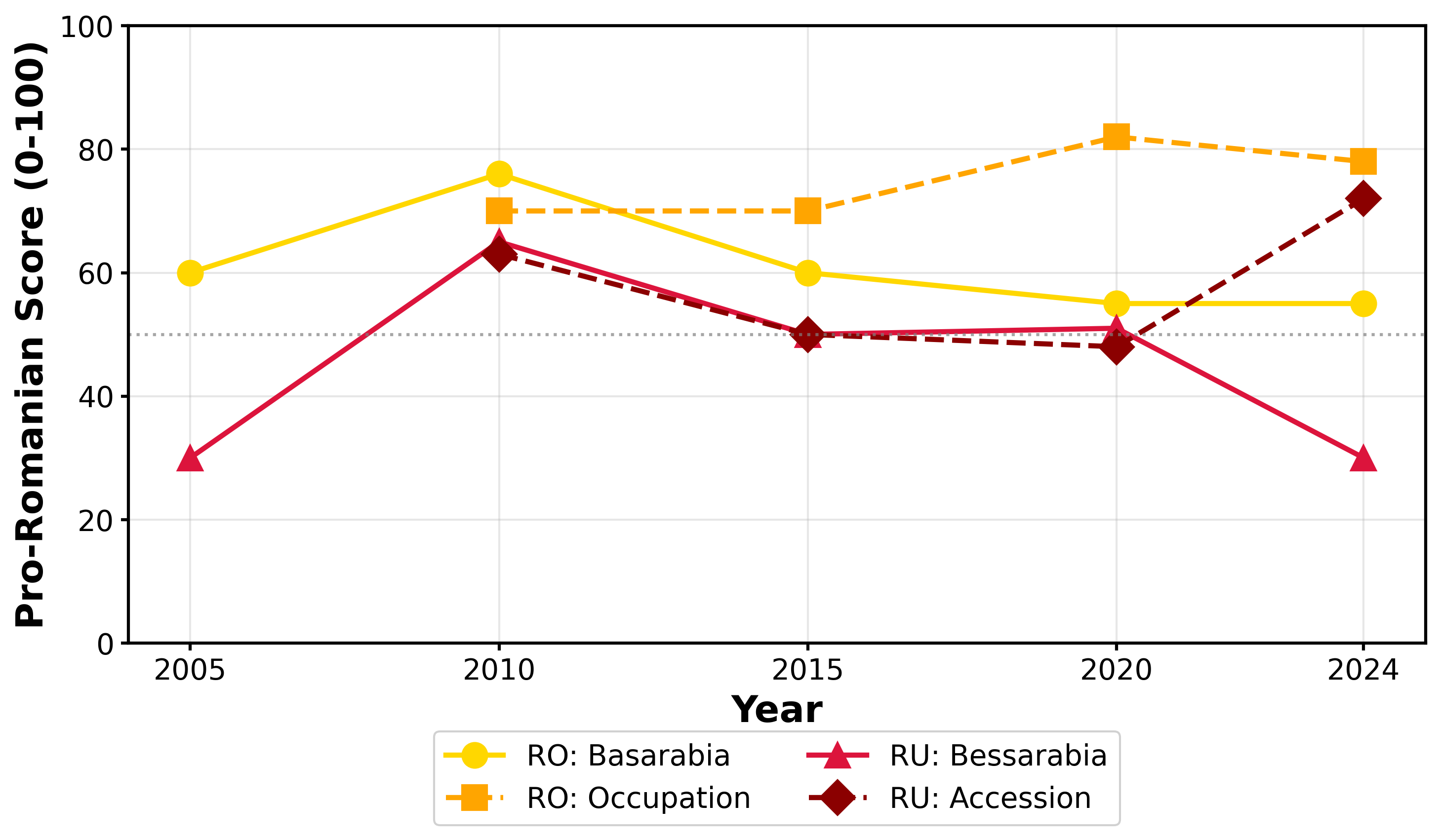}
    \includegraphics[width=0.48\textwidth]{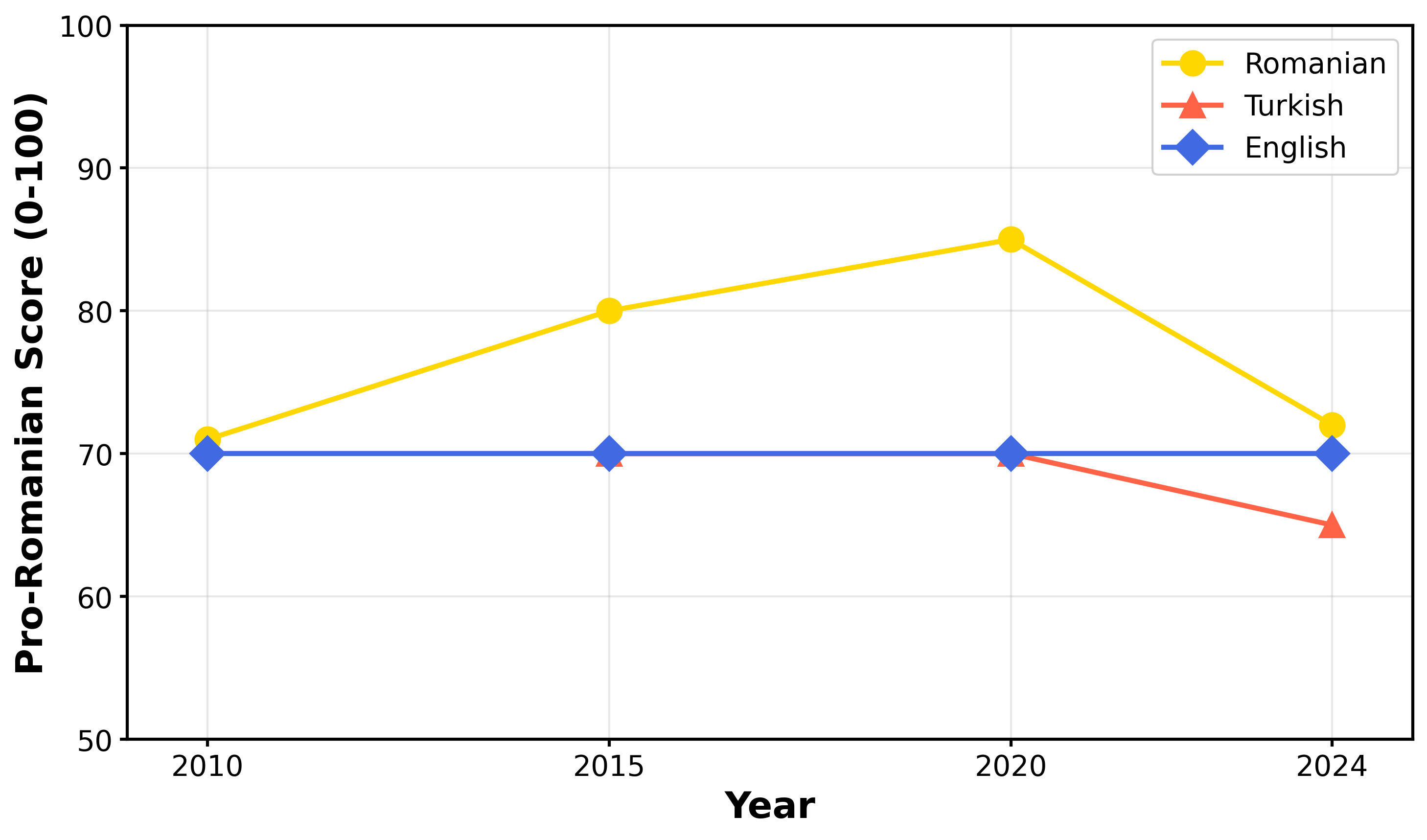}
    \caption{Evolution of the "Pro-National" score over time. Left: Bessarabia (RO vs RU). Right: Night Attack at Târgoviște (RO vs TR vs EN).}
    \label{fig:evolution}
\end{figure*}

\textbf{Bessarabia (1940):} The Romanian article for "Basarabia" showed significant moderation, with its Pro-Romanian score dropping from a peak of 76 (2010) to 55 (2024) (Figure \ref{fig:evolution}, Left). Conversely, the Russian article was highly volatile. After fluctuating between Pro-Soviet (30) and balanced (65) stances, the 2024 version of the "Accession" article spiked to a Pro-Romanian score of 72. We speculate that this may reflect a post-2022 shift where Russian editors are becoming more critical of Soviet expansionism, though further qualitative research is needed to confirm this causal link.

\begin{figure}[h]
    \centering
    \includegraphics[width=\columnwidth]{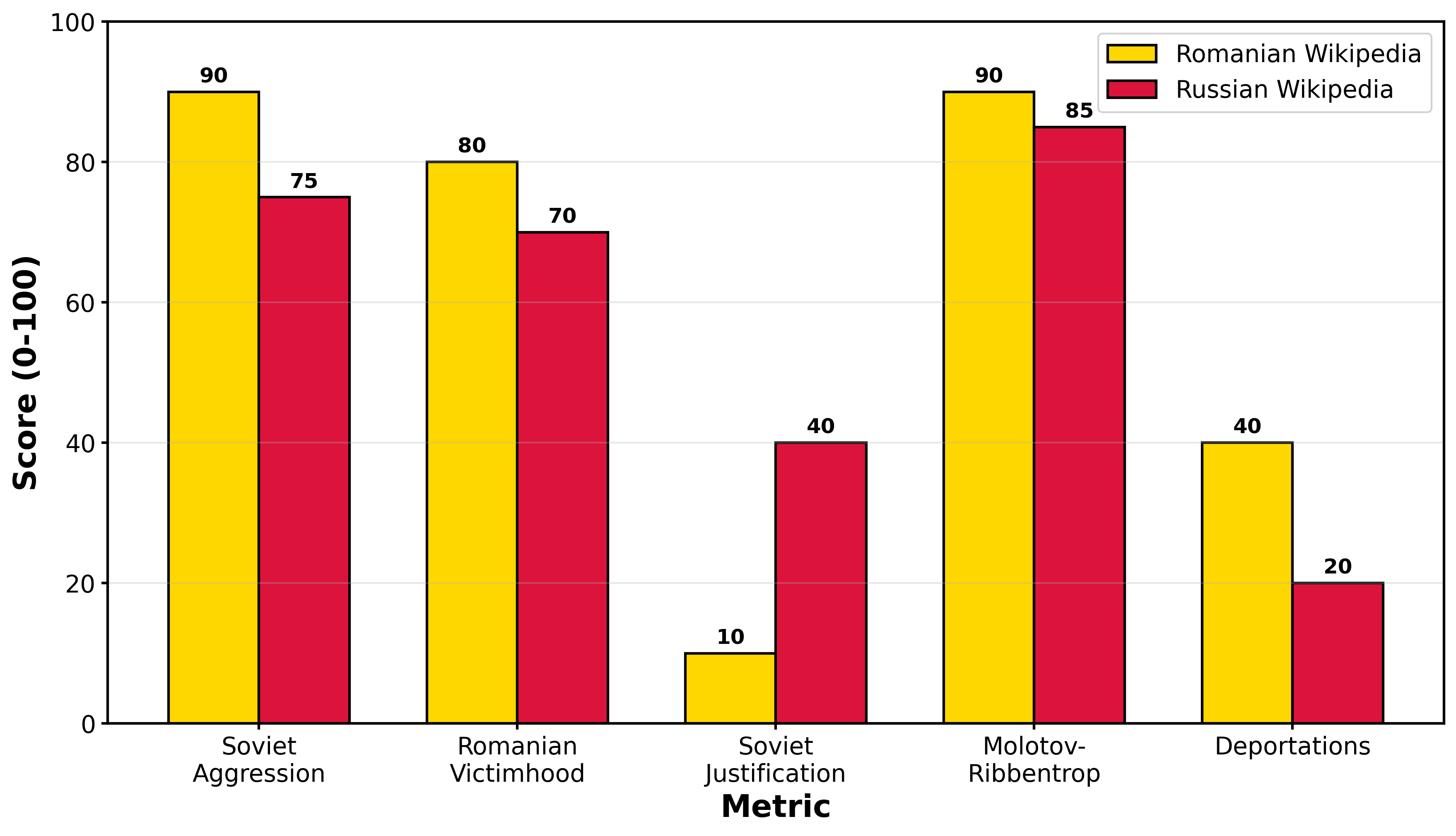}
    \caption{Detailed metric comparison for Bessarabia (2024). Note the divergence in "Soviet Justification" and "Deportations".}
    \label{fig:bessarabia_metrics}
\end{figure}

\begin{figure}[h]
    \centering
    \includegraphics[width=\columnwidth]{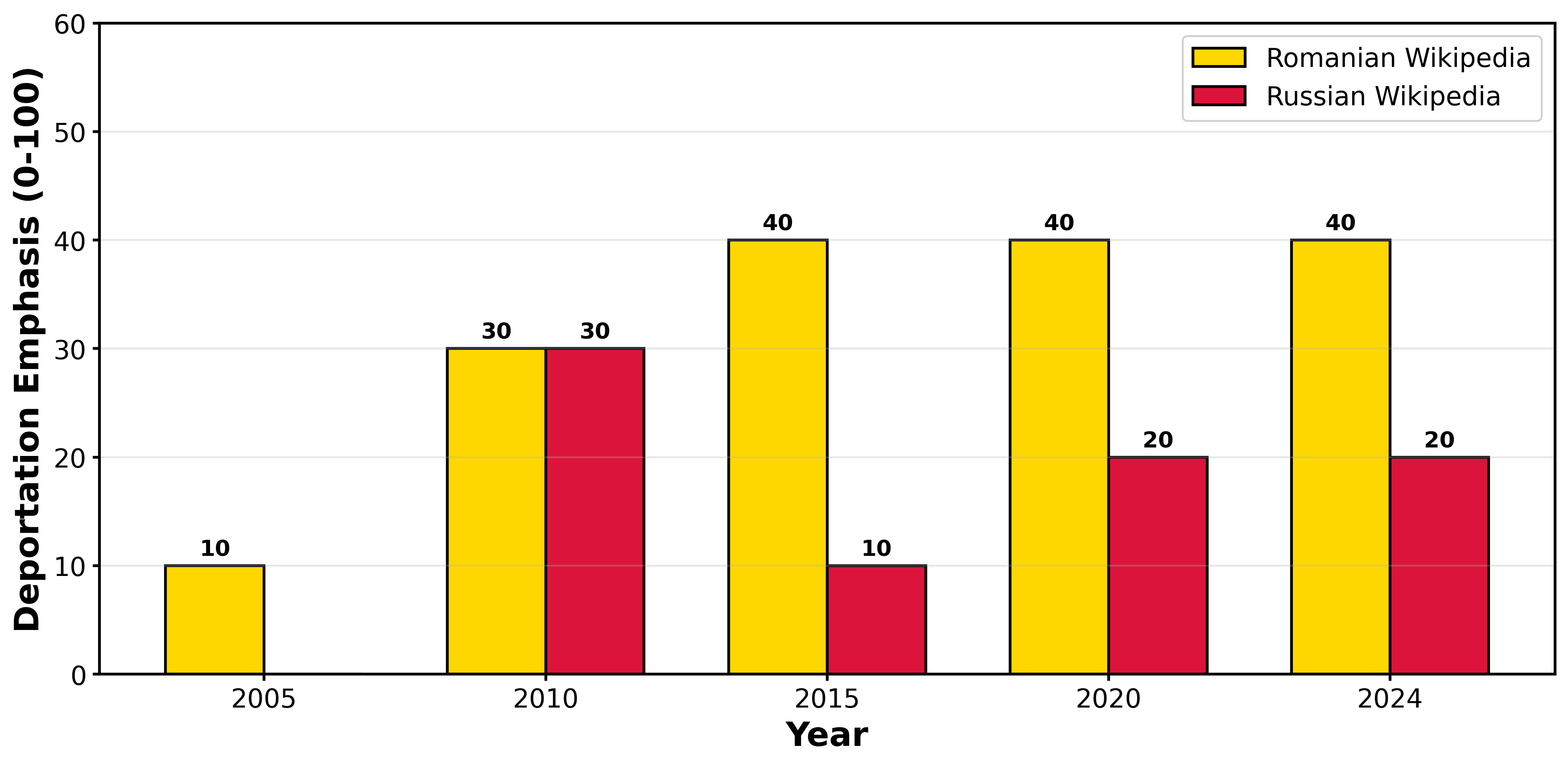}
    \caption{Deportation mentions over time. Romanian Wikipedia consistently highlights Soviet atrocities, while Russian Wikipedia largely omits them.}
    \label{fig:bessarabia_deportations}
\end{figure}

Figure \ref{fig:bessarabia_metrics} reveals the specific dimensions of this divergence. While both editions now acknowledge the Molotov-Ribbentrop Pact, they differ fundamentally on the consequences. The Romanian edition emphasizes "Deportations" (Score: 40) and "Victimhood" (Score: 80), whereas the Russian edition emphasizes "Soviet Justification" (Score: 40). Figure \ref{fig:bessarabia_deportations} tracks this "Deportation Gap" over time, showing that while Romanian mentions of atrocities have remained high, Russian mentions have been near-zero for two decades, only appearing slightly in 2024.

\textbf{Night Attack (1462):} The Romanian article peaked in nationalism in 2020 (Score: 85), portraying Vlad the Impaler as a "valiant defender", before moderating to 72 in 2024 (Figure \ref{fig:evolution}, Right). The Turkish edition showed a gradual decline in Pro-Romanian sentiment (70 $\rightarrow$ 65) and consistently emphasized Vlad's cruelty (Score: 80) far more than the Romanian edition (Score: 60). The English edition remained remarkably stable at a score of 70 throughout the 14-year period.

\begin{figure}[h]
    \centering
    \includegraphics[width=\columnwidth]{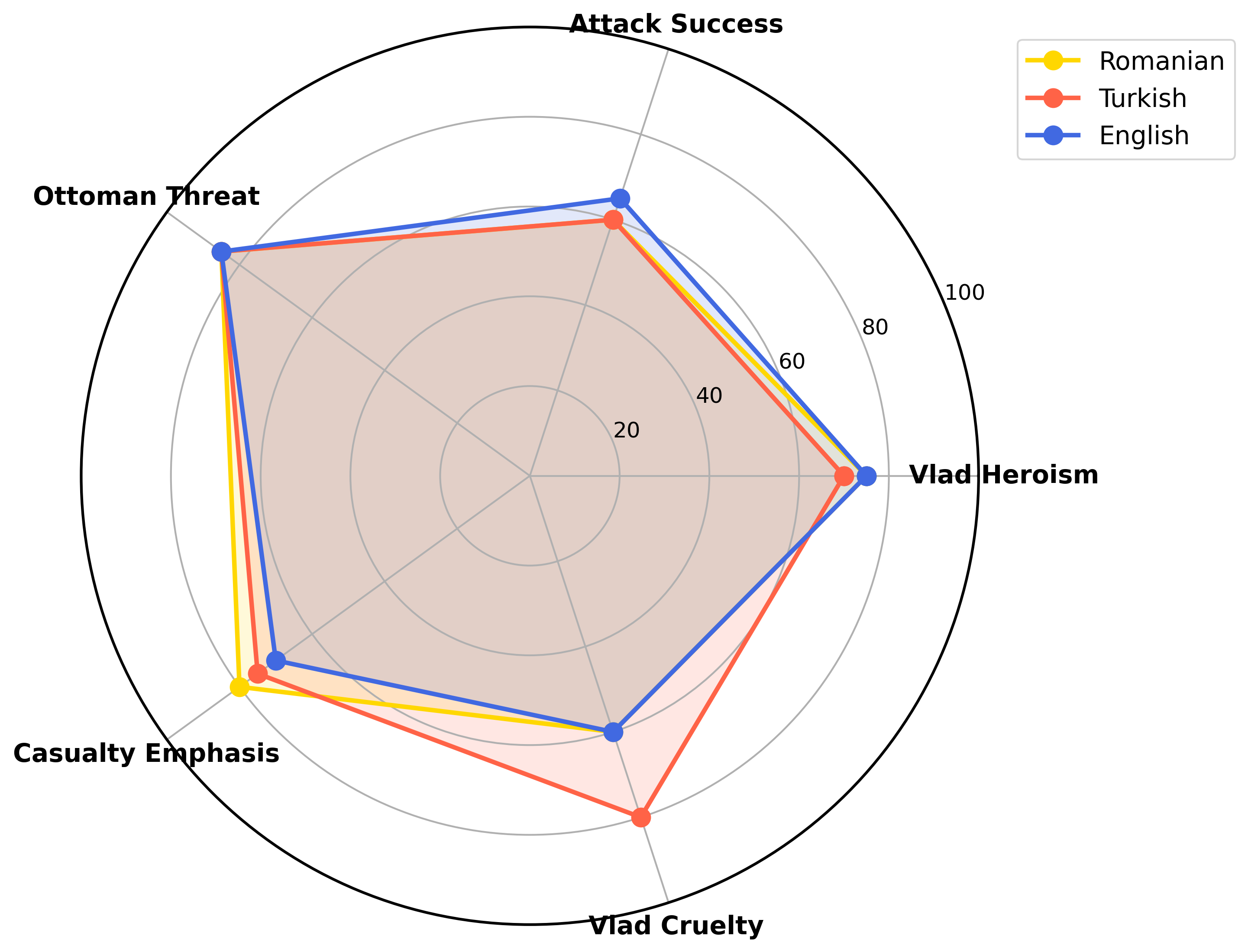}
    \caption{Radar chart of narrative metrics for the Night Attack (2024). The shapes illustrate distinct narrative priorities: Turkish (Cruelty-focused), Romanian (Heroism-focused), and English (Balanced).}
    \label{fig:targoviste_radar}
\end{figure}

\begin{figure}[h]
    \centering
    \includegraphics[width=\columnwidth]{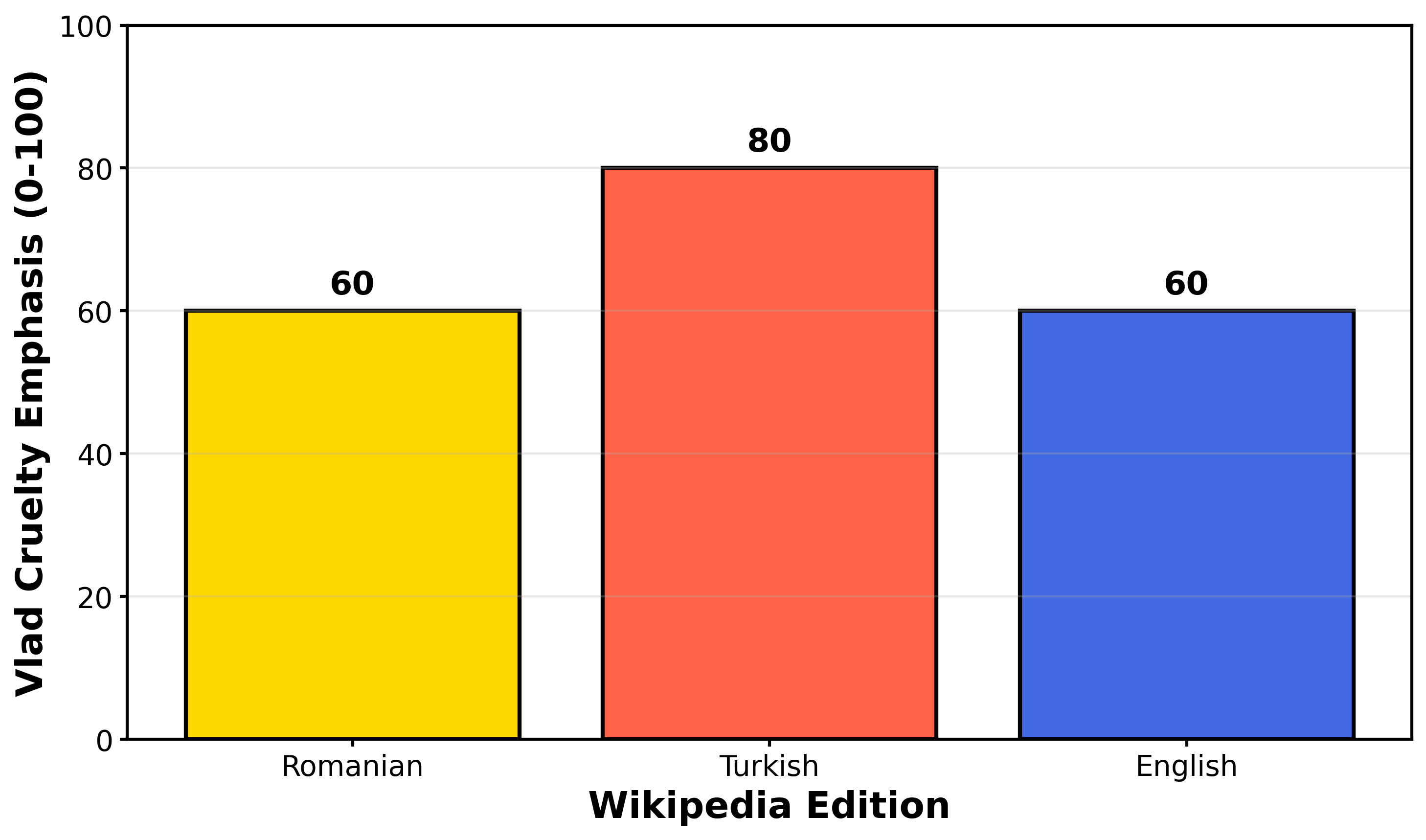}
    \caption{Comparison of "Vlad Cruelty" scores. Turkish Wikipedia consistently emphasizes Vlad's brutality much more than the Romanian edition.}
    \label{fig:targoviste_cruelty}
\end{figure}

The shape of these narratives is visualized in Figure \ref{fig:targoviste_radar}. The Turkish narrative is defined by high "Cruelty" and "Ottoman Threat" scores but lower "Heroism". The Romanian narrative is the inverse. Figure \ref{fig:targoviste_cruelty} isolates the "Cruelty Gap", showing a persistent 15-20 point difference between Turkish and Romanian portrayals of Vlad's tactics.

\subsection{Experiment 3: Peace-Maker LLM}
We evaluated four prompting strategies for generating neutral summaries of the Night Attack, using conflicting claims extracted from Romanian and Turkish articles.

\begin{figure}[h]
    \centering
    \includegraphics[width=\columnwidth]{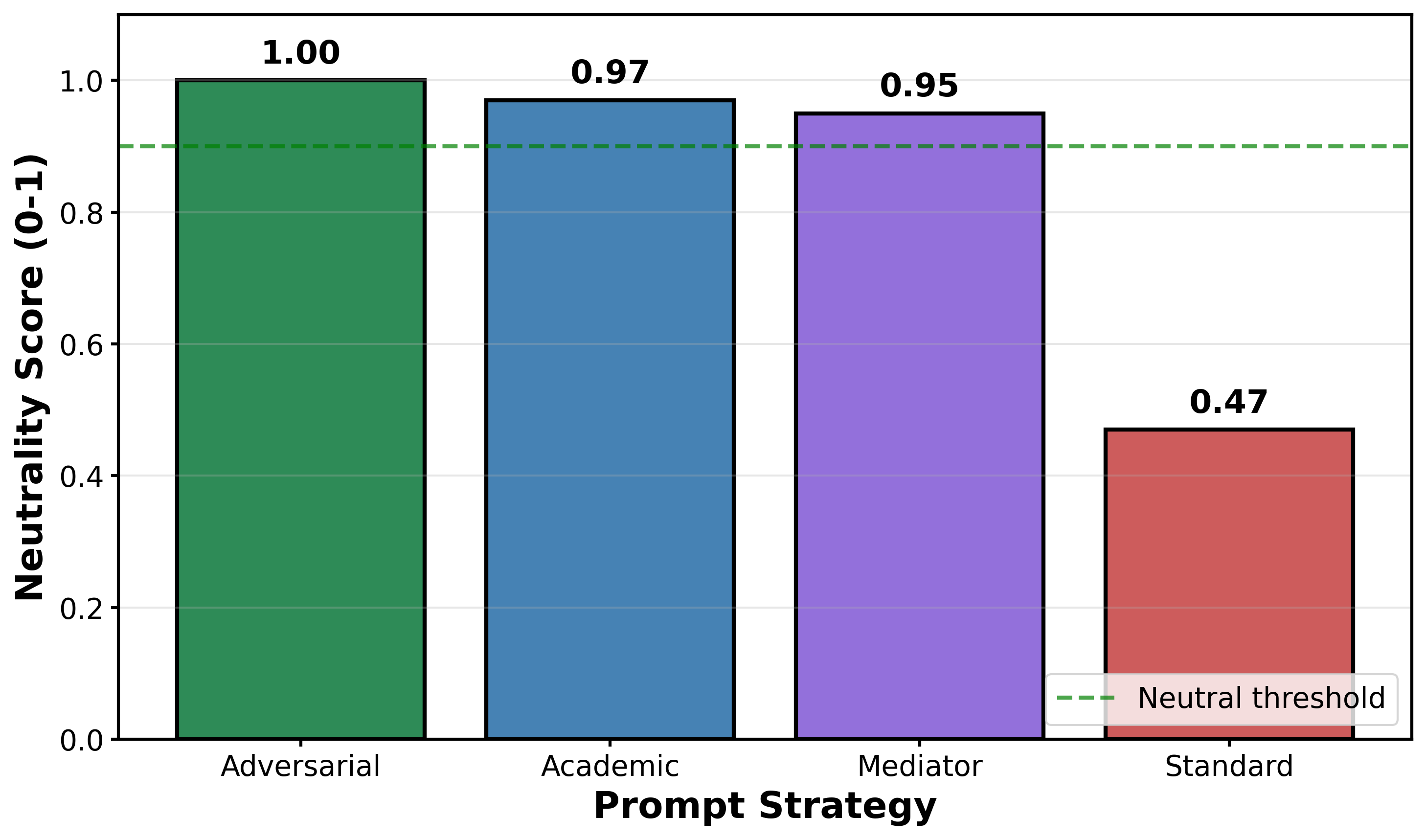}
    \caption{Neutrality scores (0-1) for different prompting strategies. The Adversarial prompt achieves perfect neutrality.}
    \label{fig:peacemaker}
\end{figure}

\begin{figure}[h]
    \centering
    \includegraphics[width=\columnwidth]{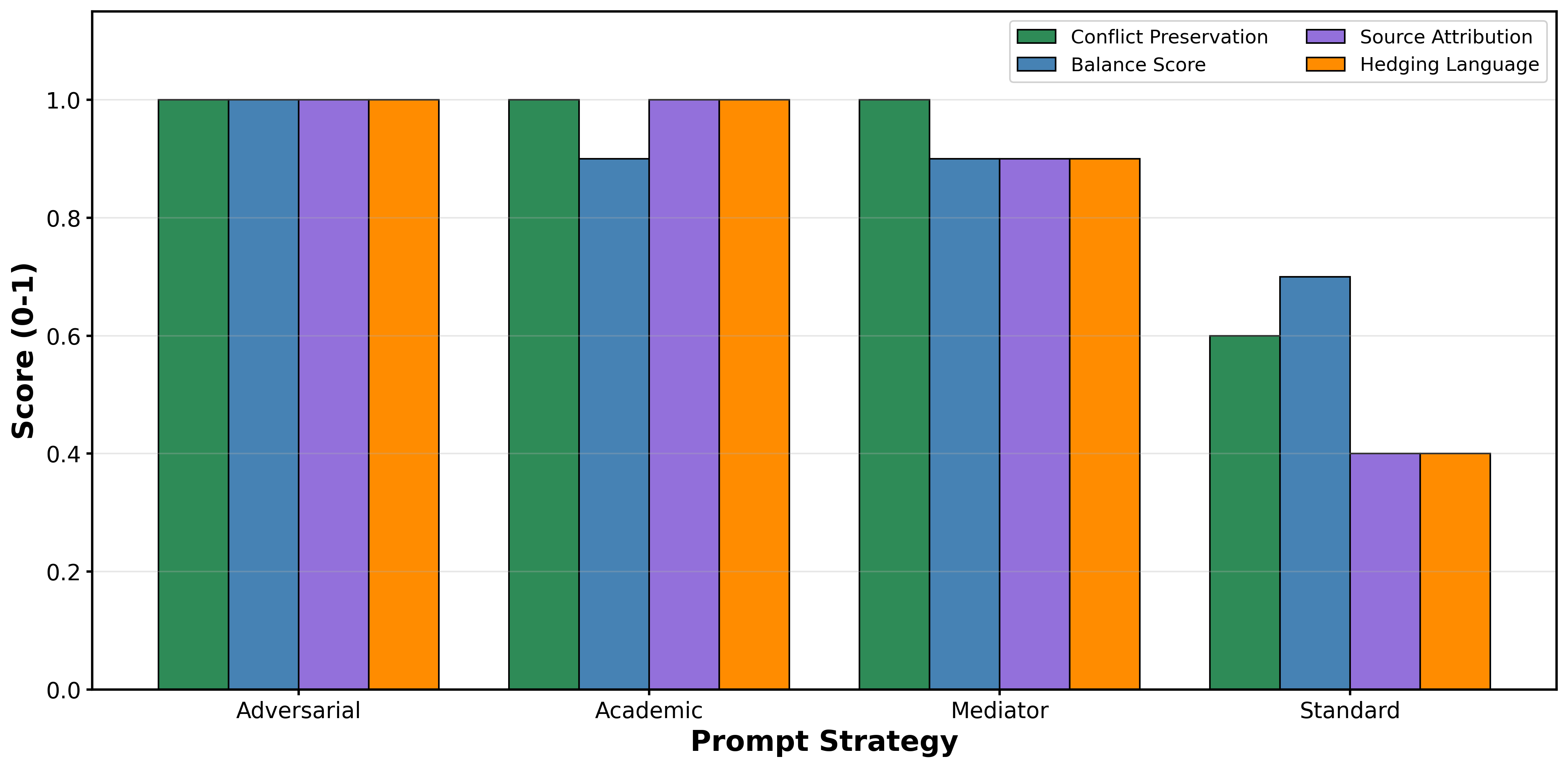}
    \caption{Detailed evaluation of generated summaries. The Adversarial prompt excels not just in Neutrality, but in Source Attribution and Conflict Preservation.}
    \label{fig:peacemaker_details}
\end{figure}

\begin{table}[h]
\centering
\begin{tabular}{lcc}
\hline
\textbf{Prompt Strategy} & \textbf{Neutrality} & \textbf{Conflict Pres.} \\
\hline
Standard & 0.47 & 0.60 \\
Mediator & 0.95 & 1.00 \\
Academic & 0.97 & 1.00 \\
\textbf{Adversarial} & \textbf{1.00} & \textbf{1.00} \\
\hline
\end{tabular}
\caption{Performance of prompting strategies. Neutrality and Conflict Preservation are scored 0-1.}
\label{tab:peacemaker}
\end{table}

As shown in Table \ref{tab:peacemaker} and Figure \ref{fig:peacemaker}, the \textit{Standard} prompt failed (Neutrality: 0.47) because the LLM attempted to resolve the conflict, often choosing one side's version of events (e.g., regarding the attack's success). The \textit{Adversarial} prompt, which explicitly instructed the model to \textit{preserve} conflict, achieved perfect scores. Figure \ref{fig:peacemaker_details} breaks this down further, showing that the Adversarial prompt's success stems from its high "Source Attribution" score—it explicitly attributes disputed claims ("Romanian sources state X..".), avoiding the trap of hallucinated consensus.

\section{Discussion}
\label{sec:discussion}

Our findings challenge the assumption that Wikipedia functions as a unified global encyclopedia. Instead, it operates as a federation of distinct epistemic communities, each validating knowledge through its own national lens.

\subsection{The Illusion of Neutrality}
The "Citation Isolation" we observed in the Posada case study shows the challenge of achieving neutrality. An article can adhere perfectly to NPOV guidelines, using neutral language and avoiding editorializing, while still presenting a heavily biased narrative simply by selecting sources from a single national tradition. This "bibliography bias" is invisible to standard NLP tools that focus on sentiment analysis, but it is the primary driver of narrative divergence in historical topics. 

\subsection{Narrative Evolution and Geopolitics}
Our temporal analysis shows that these narratives are not static. The moderation of the Romanian "Night Attack" article (from 85 to 72) suggests that community maturity and international scrutiny can temper nationalism over time. However, the volatility of the Russian "Bessarabia" article demonstrates that Wikipedia is not immune to external geopolitical shocks. The sudden 2024 shift toward a Pro-Romanian stance likely reflects a broader realignment of Russian opposition discourse following the invasion of Ukraine, where Soviet imperial history is being critically re-examined.

Bias often manifests as silence. As Figure \ref{fig:bessarabia_deportations} illustrates, the Russian Wikipedia's historical omission of Soviet deportations is a form of narrative control. The sudden appearance of these mentions in 2024 signals a potential "thaw" in this historiographical freeze. Similarly, the "Cruelty Gap" in the Targoviste narrative (Figure \ref{fig:targoviste_cruelty}) shows how national identity is constructed not just by what is celebrated (Heroism), but by what is minimized (Cruelty).

\subsection{AI as a Mediator, Not a Judge}
Standard LLM summarization fails because it mimics the human tendency to seek a single, coherent truth, effectively hallucinating a consensus where none exists. By explicitly prompting the model to be "adversarial" and preserve conflict, we force it to act as a mediator rather than a judge.

However, a distinction must be drawn between interpretive disagreements (e.g., whether a war was "justified") and factual contradictions (e.g., casualty counts). For the latter, "preserving conflict" does not imply that all claims are equally valid, but rather that the \textit{disagreement itself} is a historical fact worth reporting. When Source A claims 15,000 casualties and Source B claims 1,500, a LLM that averages these numbers fails both. A LLM that reports the discrepancy preserves the epistemic integrity of the conflict. Future systems must balance this perspectivism with rigorous fact-checking to avoid "false balance" when one perspective is demonstrably pseudohistorical. This suggests that the future of AI in education and historiography lies not in generating "objective" answers, but in synthesizing and attributing diverse perspectives.

\section{Conclusion and Future Work}
\label{sec:conclusion}
This study demonstrates that Wikipedia functions not as a unified global archive, but as a federation of distinct epistemic communities defined by "citation isolation". Our analysis of the Battle of Posada revealed that the Romanian and Hungarian editions shared only 2 out of 119 citations, constructing mutually exclusive historical realities based on non-overlapping authorities. Longitudinally, we found these narratives to be volatile and responsive to geopolitics, evidenced by the 2024 shift in the Russian framing of Bessarabia.

Methodologically, we show that standard LLM prompting fails to address this by hallucinating consensus. Instead, we propose a "Peace-Maker" pipeline using "adversarial" prompting. By explicitly instructing models to preserve conflict rather than resolve it, we achieve "perspectival balance", generating summaries that accurately attribute disagreement without enforcing a false middle ground. 

We propose several directions for future research:
\begin{itemize}
    \item \textbf{Baseline Comparison:} Compare these contested events to uncontested historical topics to establish a baseline for citation overlap and narrative divergence. This would verify whether the "citation isolation" we observed is specific to conflict or a general feature of Wikipedia's language editions.
    \item \textbf{Geographic and Domain Expansion:} While our case studies are well-chosen for their contentiousness, they are limited to Eastern European history. Future research should investigate if "Citation Isolation" holds true for global topics (e.g., WWII in the Pacific or Colonialism in the Americas) or scientific controversies.
\end{itemize}

\section*{Limitations}
Our study relies on human annotators and LLMs. LLM-based classification, while calibrated, may still contain inherent biases. We focused on three specific events in Eastern European history; results may vary for other regions. Our validation methodology presents a potential circularity: annotators assigned by language (Romanian/English vs. Hungarian) were native speakers whose historiographical perspectives may reflect the national biases we sought to measure, rather than providing an independent benchmark. Annotator neutrality is hard to verify.

\section*{Ethics Statement}

This study utilizes publicly available data under the Creative Commons Attribution-ShareAlike 4.0 license. To validate our LLM-based classifications on sensitive historical topics, we used two human annotators. Strict privacy protocols were maintained to preserve their anonymity. We emphasize that our computational metrics quantify "perspectival balance" and are not intended to adjudicate historical truth or resolve geopolitical disputes.

\section*{Acknowledgments}

This research is supported by:
\begin{itemize}
    \item the project “Romanian Hub for Artificial Intelligence - HRIA”, Smart Growth, Digitization and Financial Instruments Program, 2021-2027, MySMIS no. 351416;
    \item a grant of the Ministry of Research, Innovation and Digitization, CNCS - UEFISCDI, project SIROLA, number PN-IV-P1-PCE-2023-1701, within PNCDI IV;
\end{itemize}

\clearpage
\appendix

\section{Granular Narrative Scoring Prompt}
\label{app:prompt}

The following prompt was used for temporal analysis of the Bessarabia and Târgoviște articles:

\begin{quote}
\small
\texttt{Analyze the following Wikipedia article excerpt about Bessarabia/Soviet occupation.}\\
\texttt{ARTICLE: \{article\}}\\
\texttt{LANGUAGE: \{lang\} (Romanian or Russian)}\\
\texttt{YEAR: \{year\}}\\
\texttt{TEXT:}\\
\texttt{\{text\}}\\
\texttt{---}\\
\texttt{Analyze the NARRATIVE STANCE of this article. Consider:}\\
\texttt{1. **Overall Stance**: Is the article Pro-Romanian, Pro-Russian, Neutral, or Mixed?}\\
\texttt{   - Pro-Romanian: Portrays Soviet actions as occupation, theft, aggression}\\
\texttt{   - Pro-Russian: Portrays Soviet actions as liberation, reunification, protection}\\
\texttt{   - Neutral: Balanced presentation of facts without emotional framing}\\
\texttt{   - Mixed: Contains elements of both perspectives}\\
\texttt{2. **Key Themes**: What are the main themes/topics discussed? (list 3-5)}\\
\texttt{3. **Emotional Tone**: What is the emotional framing?}\\
\texttt{   - Accusatory (blaming the other side)}\\
\texttt{   - Neutral (factual, academic)}\\
\texttt{   - Defensive (justifying actions)}\\
\texttt{   - Victimization (emphasizing suffering)}\\
\texttt{4. **Key Terminology**: What loaded terms are used? For example:}\\
\texttt{   - "occupation" vs "liberation" vs "annexation" vs "reunification"}\\
\texttt{   - "ultimatum" vs "agreement" vs "request"}\\
\texttt{   - How is the USSR/Romania described?}\\
\texttt{Respond in this EXACT JSON format:}\\
\texttt{\{}\\
\texttt{    "overall\_stance": "Pro-RO" or "Pro-RU" or "Neutral" or "Mixed",}\\
\texttt{    "stance\_confidence": 0.0 to 1.0,}\\
\texttt{    "key\_themes": ["theme1", "theme2", "theme3"],}\\
\texttt{    "emotional\_tone": "Accusatory" or "Neutral" or "Defensive" or "Victimization",}\\
\texttt{    "terminology": \{}\\
\texttt{        "event\_name": "what the event is called",}\\
\texttt{        "soviet\_actions": "how Soviet actions are described",}\\
\texttt{        "key\_loaded\_terms": "any loaded/biased terms used"}\\
\texttt{    \},}\\
\texttt{    "reasoning": "Brief explanation of why you classified it this way"}\\
\texttt{\}}
\end{quote}

\bibliography{custom}

\end{document}